\documentclass[a4paper,fleqn]{cas-dc}

\usepackage[numbers]{natbib}
\usepackage{adjustbox} 
\usepackage{multirow}
\usepackage{graphicx}
\usepackage{lscape}
\usepackage{threeparttable}
\usepackage{tablefootnote}
\usepackage{footnote}
\usepackage{subfig}
\usepackage{placeins}

\usepackage{array}
\usepackage{makecell}

\usepackage{tabularx}

\begin{document}
\let\WriteBookmarks\relax

\shorttitle{}    

\shortauthors{Z. Yang \& H. Li et al.}  

\title [mode = title]{Cognitive Dual-Process Planning for Autonomous Driving with Structured Scene Knowledge and Verifiable Reasoning-Action Consistency}  

\tnotemark[1] 

\tnotetext[1]{This work was supported by the Key Research and Development Program of Shanxi Province (No. 202202070301005).}

%

\author[1,2]{Zhongyao Yang}

\fnmark[1]

\ead{3220240509@bit.edu.cn}


\credit{Conceptualization, Methodology, Validation, Investigation, Writing -- original draft, Writing -- review \& editing, Visualization}

\affiliation[1]{organization={School of Mechanical Engineering, Beijing Institute of Technology},
            city={Beijing},
            postcode={100081}, 
            state={Beijing},
            country={China}}

\affiliation[2]{organization={National Engineering Research Center of Electric Vehicles, Beijing Institute of Technology},
            city={Beijing},
            postcode={100081}, 
            state={Beijing},
            country={China}}

\affiliation[3]{organization={School of Mechanical Engineering, Southeast University},
            city={Nanjing},
            postcode={211189},
            state={Jiangsu},
            country={China}}

\author[1,2]{Haoyu Li}


\fnmark[1]

\ead{rabbit.yujixyz@gmail.com}


\credit{Conceptualization, Methodology, Investigation, Writing -- review \& editing, Visualization}

\author[1,2]{Yu Yan}


\ead{3220253023@bit.edu.cn}


\credit{Methodology, Investigation}

\author[1,2]{Zhuangxuan Yu}


\ead{3220253022@bit.edu.cn}


\credit{Methodology, Investigation}

\author[3]{Jiangfeng Nan}


\ead{njf@seu.edu.cn}


\credit{Methodology, Investigation}

\author[1,2]{Jinrui Nan}

\cormark[1]


\ead{nanjinrui@bit.edu.cn}


\credit{Conceptualization, Supervision, Project administration, Funding acquisition}

\cortext[1]{Corresponding author}

\fntext[1]{These authors contributed equally.}


\begin{abstract}
High-level planning for autonomous driving is a knowledge-intensive engineering decision task that requires accurate scene understanding, timely inference, and internally consistent action selection. Vision-language models (VLMs) can make intermediate reasoning explicit, but their use in deployed planners is constrained by costly structured supervision, unnecessary reasoning in routine scenes, and possible inconsistencies between generated rationales and driving actions. We present a cognitive dual-process planning framework that represents planning-relevant scene knowledge in a machine-parsable structured chain-of-thought (S-CoT) schema. An automated data engine integrates perception foundation models, critical-path filtering, and an expert VLM to generate S-CoT supervision without manual annotation of individual rationales. A lightweight visual Arbiter estimates scene complexity from multilevel vision-encoder features before language decoding and routes each input to either fast meta-action prediction or slow structured reasoning. For slow-path outputs, a deterministic rule-based validator checks whether the parsed S-CoT fields are consistent with the final meta-action and provides verifiable rewards for Group Relative Policy Optimization (GRPO). In a 195-scene manual audit, the generated annotations achieve 91.8\% CoT accuracy and a 98.5\% Logical Consistency Score (LCS). On 574 manually verified NAVSIM test samples, the planner achieves 80.14\% planning accuracy and 97.20\% LCS while reducing average latency by 17.39\% relative to applying slow reasoning to every scene. Evaluation on external long-tail subsets further identifies conditions under which routing and planning performance degrade. Together, these results show how explicit scene knowledge can be operationalized through adaptive reasoning and rule-based verification to support high-level VLM planning decisions.
\end{abstract}







\maketitle


\section{Introduction}
\label{sec:intro}

High-level planning for autonomous driving converts visual observations, ego-vehicle states, and navigation instructions into driving decisions subject to latency and safety constraints. End-to-end models reduce hand-engineered interfaces by learning this mapping directly from data \cite{bojarski2016end, codevilla2018end, hu2023planning, weng2024drive}. Their performance can nevertheless degrade in long-tail and out-of-distribution scenes, where a decision depends on relations among traffic participants, road structure, signals, and route constraints \cite{zhou2024vision, chen2024end}. These challenges are amplified by adverse weather or sensor degradation, which can corrupt object- and road-structure cues before they reach downstream planning modules \cite{li2026robo}. High-level planning is therefore more than an action-prediction problem: it requires explicit representations of planning-relevant scene knowledge, adaptive allocation of reasoning resources, and consistency checks between inferred scene factors and selected actions.

Vision-language models (VLMs) can support this formulation because they combine visual representations with language-based task knowledge \cite{bai2025qwen2, liu2023visual, chen2024internvl}. Their language decoders can generate explicit chain-of-thought (CoT) rationales that link scene factors to driving decisions \cite{xu2024drivegpt4, xing2025openemma, wang2024drivecot}. However, studies of large models for intelligent transportation identify task alignment and inference cost as central deployment constraints \cite{gan2024large}. Lightweight driving VLMs likewise treat model complexity and response time as first-order design requirements \cite{gan2026performance}. Nevertheless, three interrelated challenges remain when VLM reasoning is used for high-level planning.

\textbf{1) Structured supervision at scale.} Free-form rationales are costly to annotate and difficult to convert into reusable or verifiable knowledge. Explicit process-knowledge representations have been shown to improve the traceability and reuse of decision information in complex engineered systems \cite{wang2021process}. However, high-level driving still lacks a machine-parsable schema that explicitly connects environmental conditions, critical objects, lane and route constraints, decision intentions, and final actions.

\textbf{2) Reasoning-resource allocation.} Applying full CoT reasoning to every scene increases latency in routine situations. Fast--slow architectures and self-routing methods seek to reduce this overhead \cite{tian2024drivevlm, xu2025towards, qian2024fasionad, zhou2025autovla, luo2025adathinkdrive}. However, routing within an autoregressive language model often couples the routing decision to text generation and offers limited control over the false-negative rate for complex scenes. These limitations motivate a separate routing mechanism that operates directly on intermediate visual features and exposes a continuous, threshold-controlled operating point.

\textbf{3) Verifiable reasoning--action consistency.} A generated rationale may identify a relevant hazard yet produce an incompatible driving action. Standard supervised fine-tuning learns output sequences but does not itself enforce consistency among parsed scene factors, reasoning intentions, and actions under a shared set of traffic constraints. Existing language--action alignment methods improve task coupling \cite{wen2023dilu, renz2025simlingo}; we complement them with a deterministic signal that verifies structured reasoning--action consistency.

We address these problems with a cognitive dual-process framework for high-level VLM planning. An automated data engine combines perception foundation models, critical-path filtering, and an expert VLM to encode planning-relevant scene knowledge in a machine-parsable structured CoT (S-CoT) schema. A decoupled visual Arbiter estimates scene complexity from multilevel vision-encoder features and routes each input to fast meta-action prediction or slow structured reasoning. For slow-path outputs, a rule-based validator parses the S-CoT fields and checks their consistency with the final meta-action. These structured outputs enable the computation of base-task, logical-consistency, and risk-aversion rewards for Group Relative Policy Optimization (GRPO) within a Reinforcement Learning with Verifiable Rewards (RLVR) procedure.

We train and validate the framework on a high-level planning dataset sampled from the NAVSIM trainval split and evaluate it on a manually verified subset of the NAVSIM test split \cite{caesar2021nuplan, dauner2024navsim}. The automated annotations reach 91.8\% CoT accuracy and 98.5\% Logical Consistency Score (LCS). The resulting planner achieves 80.14\% planning accuracy and 97.20\% LCS while reducing average latency by 17.39\% relative to static slow reasoning. Targeted external long-tail subsets probe behavior beyond the NAVSIM-derived distribution and reveal reduced robustness in low-visibility and traffic-sign reasoning scenarios. The main contributions are summarized as follows.

\begin{itemize}
  \item We introduce an S-CoT knowledge representation that organizes scene evidence, lane and route constraints, reasoning intentions, and meta-actions into machine-parsable fields, together with an automated data engine that generates grounded supervision without manual annotation of individual rationales.
  
 \item We develop a decoupled visual Arbiter that estimates a continuous scene-complexity score from multilevel vision-encoder features, enabling threshold-controlled routing between direct meta-action prediction and structured slow reasoning.
  
  \item We integrate a deterministic rule-based validator with GRPO to train the slow reasoning path using base-task, logical-consistency, and risk-aversion rewards that target inconsistencies between parsed S-CoT fields and discrete meta-actions.
\end{itemize}

\begin{figure*}[pos=t]
  \centering
  \includegraphics[width=0.9\textwidth]{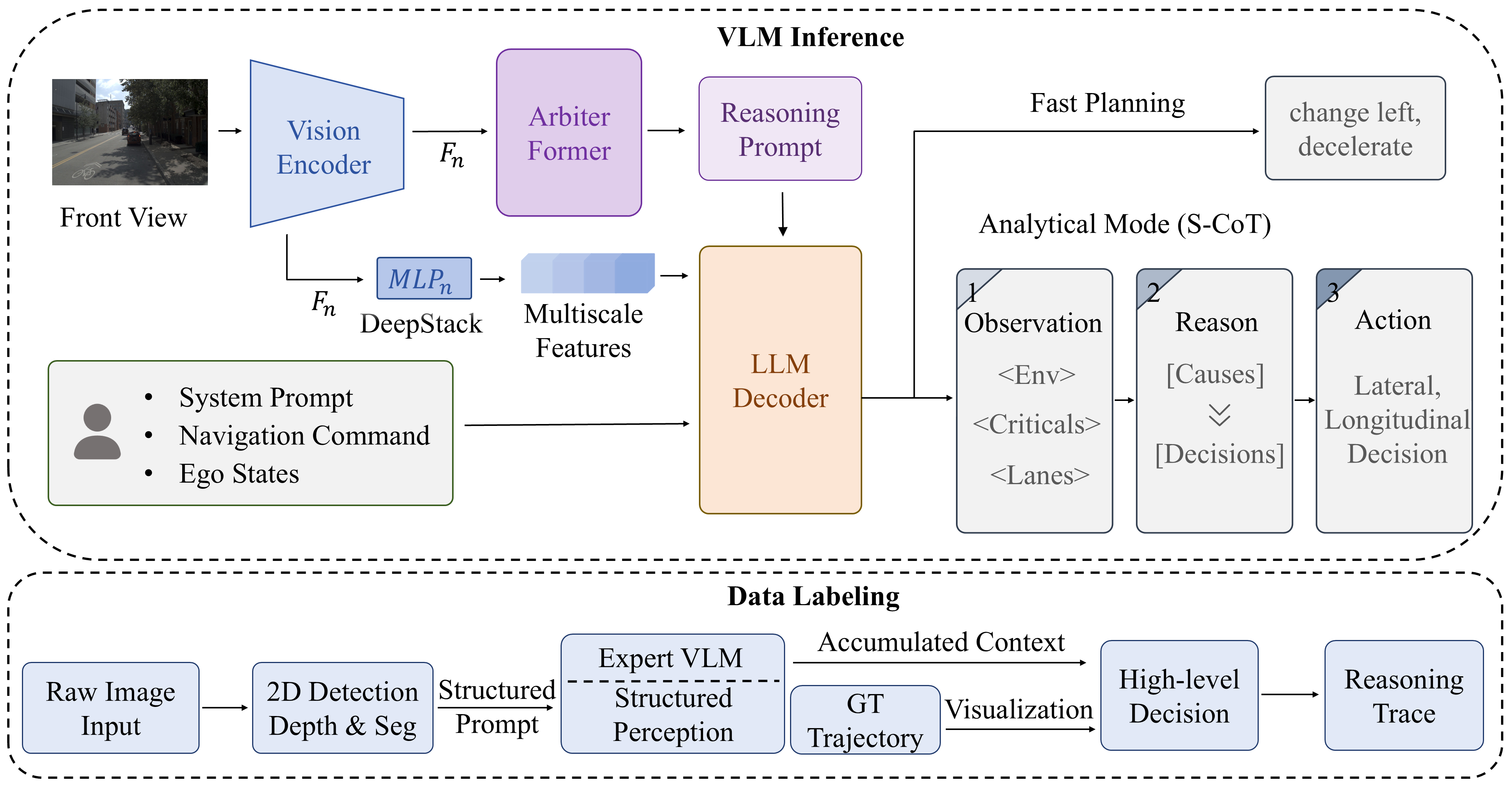}
  \caption{
     Overview of the proposed dual-process architecture. The planner takes front-view images, navigation commands, and ego-vehicle states as input. A lightweight Arbiter estimates scene complexity and routes each input to fast meta-action prediction or slow structured reasoning. The automated data engine provides structured chain-of-thought (S-CoT) supervision, while rule-based verification supplies rewards that target consistency between the parsed reasoning fields and the final meta-action.
  }
  \label{f1} 
\end{figure*}


\section{Related Work}
\subsection{High-Level Planning for Autonomous Driving}
Classical autonomous driving systems separate perception, prediction, and planning into modular components \cite{chu2025survey, li2025delving}. This decomposition provides explicit interfaces, but errors can propagate across them. End-to-end planners instead learn a direct mapping from sensor observations to trajectories or driving decisions \cite{hu2023planning, shao2023reasonnet, weng2024drive, jiang2023vad, li2024hydra, sun2025sparsedrive, liao2025diffusiondrive}. Many of these methods optimize geometric trajectories and closed-loop driving scores. Their latent representations do not necessarily expose how scene knowledge supports a high-level decision, making knowledge--action consistency difficult to inspect directly. Related surveys identify task-alignment and deployment constraints when large language models are introduced into the intelligent-transportation planning stack \cite{gan2024large}.

VLM-based planners add semantic scene interpretation and language-conditioned reasoning. DriveGPT4 generates natural-language explanations alongside vehicle control signals \cite{xu2024drivegpt4}, while Senna separates VLM-based high-level planning from low-level control \cite{jiang2024senna}. OpenDriveVLA unifies multimodal reasoning and action generation within a vision--language--action model \cite{zhou2025opendrivevla}. Reason2Drive focuses on evaluating the causal structure of driving explanations rather than executable planning \cite{nie2024reason2drive}. Together, these methods connect perception, textual reasoning, and control at different levels of the planning stack. In contrast, the present study predicts discrete high-level meta-actions and evaluates whether machine-parsable scene knowledge is consistent with the selected actions.

\subsection{Structured and Adaptive VLM Reasoning}
Recent driving VLMs further introduce structured intermediate reasoning to improve action prediction and long-tail generalization
\cite{wang2025alpamayo}. Chain-of-thought (CoT) methods generate intermediate steps before a final output, with extensions based on self-consistency, search, and program-aided reasoning \cite{wei2022chain, wang2022self, yao2023tree, gao2023pal}. In autonomous driving, VLM-AD uses teacher-generated free-form reasoning for supervision \cite{xu2024vlm}, and Think-Driver produces textual explanations linking scene understanding to decisions \cite{zhang2024think}. Free-form rationales are readable but difficult to translate into machine-checkable planning constraints. Structured alternatives narrow the output space: SOLVE incorporates intermediate reasoning into trajectory prediction \cite{chen2025solve}, while DriveRX organizes reasoning through cross-task visual question answering \cite{diao2025driverx}. These approaches connect structured reasoning with planning, but their focus is the reasoning structure itself rather than pre-decoding routing between direct prediction and structured reasoning.

Adaptive reasoning seeks to reduce the computational cost of invoking full CoT for every scene. DriveVLM-Dual combines a VLM reasoning branch with a conventional driving stack \cite{tian2024drivevlm}, and FASIONAD coordinates fast and slow systems through score distributions and feedback \cite{qian2024fasionad}. AutoVLA and AdaThinkDrive train autoregressive models to select between direct planning and CoT-assisted generation \cite{zhou2025autovla, luo2025adathinkdrive}. Model compression and visual-token refinement provide complementary approaches to reducing VLM latency \cite{gan2026performance}. These methods typically couple routing to branch fusion, generated tokens, or the planning policy. In contrast, we formulate routing as a separately supervised visual decision made before language decoding and evaluate its operating point in terms of false-negative rate (FNR), latency, and planning accuracy.

\subsection{Automated Supervision and Verifiable Policy Alignment}
Recent studies use foundation models to generate driving data and supervision. GAIA-2 generates controllable multi-view driving scenes \cite{russell2025gaia}, whereas OmniDrive uses counterfactual synthetic annotations to connect 3D perception, reasoning, and planning \cite{wang2025omnidrive}. These pipelines can reduce dependence on manually created data or annotations, but their outputs target scene generation, visual question answering, or trajectory supervision. Our data engine instead produces a fixed schema of scene and intention fields for structured reasoning; these fields can be checked before training.

Reinforcement fine-tuning provides a second mechanism for task alignment. Proximal Policy Optimization (PPO) is widely used in Reinforcement Learning from Human Feedback (RLHF) to optimize language models with learned or human-derived reward signals \cite{schulman2017proximal, stiennon2020learning, ouyang2022training}, whereas GRPO removes the learned critic and uses relative rewards within sampled groups \cite{shao2024deepseekmath, guo2025deepseek}. In autonomous driving, AlphaDrive defines GRPO rewards for planning and reasoning \cite{jiang2025alphadrive}, whereas knowledge-transfer and teacher--student reinforcement-learning frameworks improve training safety and efficiency as the decision environment becomes increasingly complex \cite{zhou2025knowledge}. These methods mainly optimize policy performance, training efficiency, or behavioral safety. In contrast, our framework uses deterministic domain rules to verify consistency among parsed scene knowledge, reasoning intentions, and final discrete meta-actions.

\begin{figure*}[pos=t]
  \centering
  \includegraphics[width=0.9\textwidth]{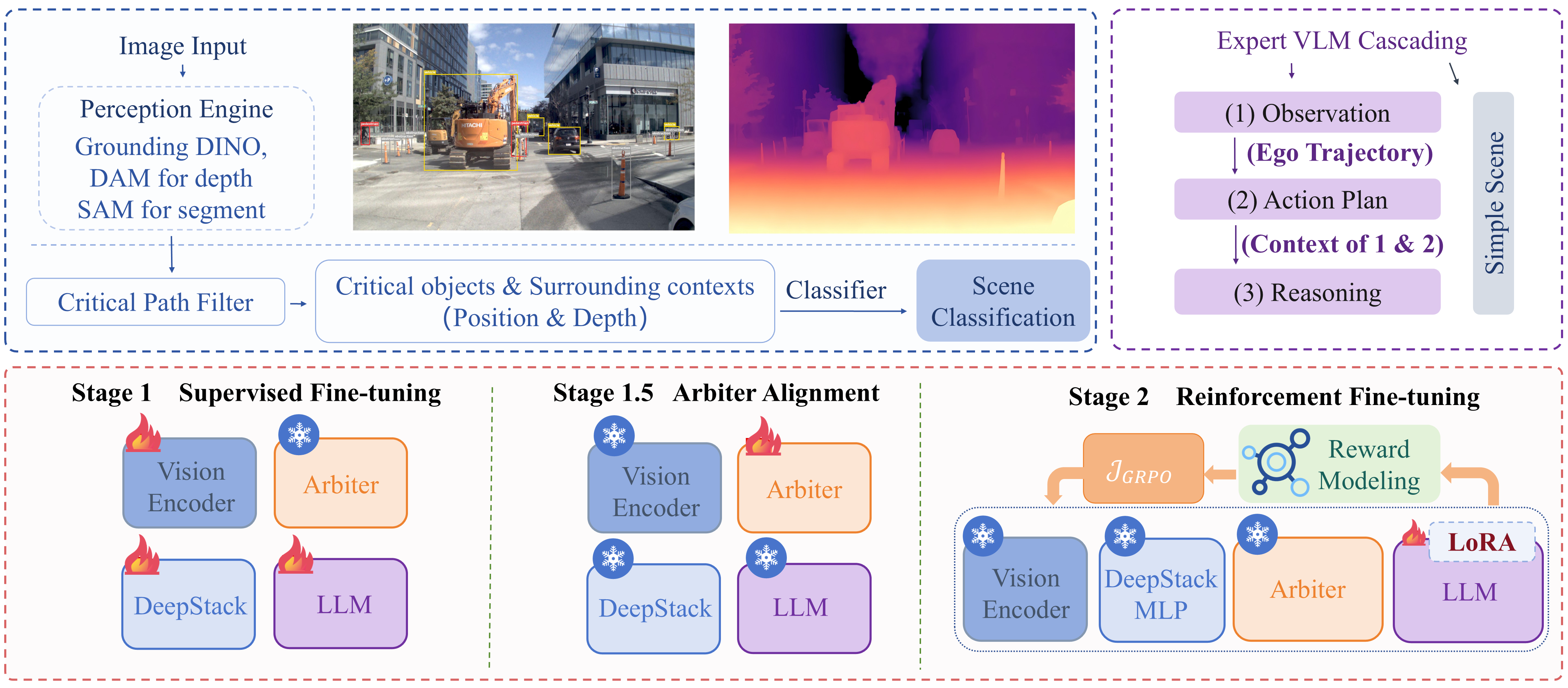}
  \caption{
    Automated data engine and hierarchical training workflow. The offline pipeline uses perception foundation models and an expert vision-language model (VLM) in sequence to generate structured chain-of-thought (S-CoT) annotations and scene-complexity labels. Training comprises a backbone supervised fine-tuning (SFT) warm-up followed by independent Arbiter calibration and Group Relative Policy Optimization (GRPO)-based slow-path optimization.
  }
  \label{f2} 
\end{figure*}


\section{Method}
\label{sec:method}

\subsection{Overview}
\label{ssec:overview}
The framework adopts Qwen3-VL as the VLM backbone \cite{bai2025qwen3} and uses DeepStack features from multiple vision-encoder layers \cite{meng2024deepstack}. As shown in Figure~\ref{f1}, the automated data engine generates S-CoT supervision and labels of scene complexity for subsequent offline training. During deployment, the visual Arbiter estimates complexity before language decoding and routes each input either to direct meta-action prediction or to structured slow reasoning. SFT establishes the two output modes, after which RLVR optimizes consistency among scene factors, reasoning intentions, and final meta-actions for slow-path samples.

\subsection{Automated Data Engine and S-CoT Generation}
\label{ssec:data_engine}
The automated data engine converts raw sensory inputs into structured S-CoT supervision organized into predefined reasoning fields, without requiring manual rationale annotation for each sample (Figure~\ref{f2}). The offline pipeline comprises perception-prior extraction, critical-path filtering, expert annotation, scene-complexity labeling, and action-anchored rationale generation.

\textbf{Perception Prior Extraction.} We use perception foundation models to obtain object-level spatial priors from each image. Grounding DINO \cite{liu2024grounding} detects traffic elements and dynamic or static objects, while the Segment Anything Model \cite{kirillov2023segment} and Depth Anything Model \cite{yang2024depth} provide instance masks and relative depth estimates, respectively. These outputs support the subsequent filtering and annotation stages.

\textbf{Critical Path Filtering.} To refine these detections, we apply VLM-based filtering to the extracted entities to reduce false positives. During offline annotation, a critical-path filter then categorizes the validated objects relative to a perspective-scaled driving corridor projected from the future ego trajectory provided by the dataset. The pixel-wise intersection between the instance masks and this spatial corridor distinguishes on-path critical objects that may affect driving decisions from peripheral context. The perception foundation models and future trajectory are used only for offline data construction and are not available to the deployed Arbiter or planner.

\textbf{Offline Expert Annotation and Complexity Labeling.} Qwen3-VL-235B-A22B serves as the offline expert annotator, whereas the deployed planner is obtained by fine-tuning Qwen3-VL-4B. This separation confines the larger model to dataset construction and prevents the target planner from generating its own reasoning supervision. Scene-complexity labels are derived from the extracted physical priors and route context. A scene is labeled as simple only when it corresponds to unobstructed cruising or stable car-following without peripheral threats. A scene is labeled as complex if at least one of three risk dimensions is present: (1) environmental risk, including traffic lights, nighttime, adverse weather, construction zones, or complex intersections; (2) density risk, including multiple nearby dynamic vehicles or pedestrians; or (3) strategic-interaction risk, in which an on-path object or lane conflict requires a non-cruising maneuver.

\textbf{Structured Knowledge Injection and Formulation.} The categorized objects are formatted as structured text (class, region, and depth) and supplied to the expert VLM, together with the visual input, through a multimodal prompt. The prompt also includes the ego-vehicle state, navigation command, critical-object category, lane availability, and relative-depth ordering. The model generates the S-CoT according to a strict template comprising environmental condition, critical objects, lane availability, action planning, and brief reasoning. We introduce discrete labels such as \texttt{uncertain} and \texttt{conditional} into the perceptual stages to train the model to represent environmental ambiguity.

\textbf{Action-Anchored Reverse-Order Annotation.}
To reduce contradictions between generated rationales and meta-action labels,
we reverse the output order during offline annotation. Unlike online inference,
offline annotation first requires the expert VLM to predict the discrete meta-action and then generate an
action-conditioned rationale while keeping the selected action fixed. Before
SFT, the resulting annotations are reordered into the standard inference
sequence, with the rationale preceding the final meta-action. This procedure uses
the action as an annotation anchor while preserving the reasoning-before-action sequence
during deployment.

\subsection{Scene Arbiter Module} 
\label{ssec:Arbiter}
The Arbiter draws on robotic arbitration architectures that select among control systems \cite{liberzon1999basic, brooks1986robust}. It predicts scene complexity directly from intermediate visual features before text generation, rather than performing routing within the autoregressive language model. Separating visual arbitration from autoregressive language generation also makes the invocation of structured reasoning independently inspectable. This design is consistent with engineering XAI principles that expose semantically meaningful intermediate information rather than relying solely on post-hoc explanations of a monolithic prediction \cite{geyer2024explainable}.

The decoupled Arbiter is a lightweight Transformer \cite{vaswani2017attention} that receives features $F_l \in \mathbb{R}^{N \times d_{\text{model}}}$ from $M=4$ ViT layers, where $d_{\text{model}}=2560$. To process these diverse visual cues, it uses $K$ learnable queries $Q_{\text{arbiter}} \in \mathbb{R}^{K \times d_{\text{model}}}$. The first query is a global complexity token $q_{\text{cls}}$, and the remaining $K-1$ queries extract localized cues. Within each Transformer block, the local queries perform cross-attention over the concatenated multilevel features to extract fine-grained, localized risks, while self-attention exchanges evidence among all queries and aggregates it into $q_{\text{cls}}$. The updated complexity token is then passed directly to the classification head:

\begin{align}
  H_{\text{cls}} &= \text{Decoder}(Q_{\text{arbiter}}, \text{Concat}(F_1, \dots, F_M)). \label{eq:H_complex}
\end{align}
The classification head converts $H_{\text{cls}}$ into a continuous scene-complexity probability:
\begin{align}
  P_{\text{complex}} &= \sigma(\text{MLP}_{\text{head}}(H_{\text{cls}})), \label{eq:P_complex}
\end{align}
where $\sigma$ denotes the sigmoid function.

The Arbiter is trained independently as a binary classifier. Because routing a complex scene incorrectly to the fast path constitutes a safety-relevant classification error, the objective includes a false-negative penalty:
\begin{align}
  \mathcal{L}_{\text{arbiter}} &= \mathcal{L}_{\text{BCE}} + w_{\text{r}} \cdot \mathcal{L}_{\text{risk}}, \\
  \mathcal{L}_{\text{risk}} &= - C_{\text{r}} \cdot \log P_{\text{complex}},
\end{align}
where $w_{\text{r}}$ weights the penalty. For complex training samples, $\mathcal{L}_{\text{risk}}$ increases as $P_{\text{complex}}$ decreases. The coefficient $C_{\text{r}}$ represents the offline-estimated consequence of skipping slow reasoning for that sample.

To avoid circular dependence between routing and planning, $C_{\text{r}}$ is computed offline with the SFT planner frozen. Let $\hat{a}_{\text{fast}}(x)$ denote the meta-action obtained by forcing the use of the fast-path prompt. The cost increases when this action disagrees with the ground-truth action or violates a validator rule:
\begin{align}
  C_{\text{r}}(x)
  &= \alpha
  \begin{cases}
    1, & \hat{a}_{\text{fast}}(x)\neq a_{\text{gt}}(x),\\
    0, & \text{otherwise},
  \end{cases}
  + \gamma\,\mathcal{V}\!\left(\hat{a}_{\text{fast}}(x),x\right),
\end{align}
where $\mathcal{V}(\hat{a}_{\text{fast}}(x),x)$ evaluates the fast-path action against the parsed scene context for sample $x$ and returns 1 if a validator rule is violated. We set $\alpha=\gamma=1$ and keep $C_{\text{r}}$ fixed during Arbiter training.

\subsection{Supervised Fine-Tuning}
\label{ssec:sft}
SFT trains the VLM backbone using separate fast- and slow-path prompts with a cross-entropy objective.

For simple scenes, the VLM is trained to output meta-actions directly:
\begin{align}
  L_{\text{fast}} &= L_{\text{CE}}(\pi_{\theta}(o | q, \text{Prompt}_{\text{fast}}), \text{Action}_{\text{GT}}) \label{eq:L_fast},
\end{align}
where $q$ is the visual input and $o$ is the meta-action token output. For complex scenes routed to the slow path, the VLM is trained to output the complete S-CoT sequence using a segment-wise weighted cross-entropy loss:
\begin{align}
  L_{\text{slow}} &= -\lambda \cdot \sum_{t=1}^{T} w_t \cdot \log(\pi_{\theta}(o_t | o_{<t}, q, \text{Prompt}_{\text{slow}})) \label{eq:L_slow},
\end{align}
where $w_t$ weights each S-CoT segment and $\lambda$ is a global balancing coefficient.

\subsection{Reinforcement Learning with Verifiable Rewards}
\label{ssec:reward}
Starting from the SFT model, RLVR applies GRPO to compare sampled slow-path outputs using task-specific verifiable rewards, without training a separate critic \cite{shao2024deepseekmath}. The policy $\pi_{\theta}$ is optimized using the composite reward:
\begin{align}
  \mathcal{R} &= \mathcal{R}_{\text{base}} + \mathcal{R}_{\text{logic}} + \mathcal{R}_{\text{risk}}.
\end{align}

\textbf{Base Task Reward.} The reward $\mathcal{R}_{\text{base}}$ scores adherence to the S-CoT format and final-action accuracy. A sigmoid length penalty discourages verbose rationales, and additional credit for correct non-conservative maneuvers, such as lane changes, reduces bias toward overly conservative actions.

\textbf{Logical Consistency Reward.} For complex scenes, $\mathcal{R}_{\text{logic}}$ rewards accurate intermediate perception to preserve the perceptual capability acquired during SFT and uses a rule-based validator to penalize reasoning--action contradictions. For example, a deterministic penalty is applied when the output identifies an immediate frontal hazard but fails to select a deceleration action.

\textbf{Risk Aversion Reward.} The reward $\mathcal{R}_{\text{risk}}$ promotes prudent model behavior in high-risk situations. It rewards outputs that recall prudent behaviors and provides additional reinforcement when uncertainty is appropriately expressed through the discrete ambiguity labels defined in the S-CoT schema.

\textbf{Deterministic Consistency Validator.} The validator parses a fixed set of S-CoT slots and applies the traffic-scene rules in Table~\ref{tab:validator_rules}. A sample passes only if no hard contradiction is detected. During RLVR, the same rules compute $\mathcal{R}_{\text{logic}}$ for sampled outputs. For the reasoning-intention check, we parse a small set of constrained intention verbs used during data generation and compare them with the final meta-action.

\begin{table*}[pos=tbp]
  \centering
  \caption{Rule-based conditions used by the validator during RLVR training and LCS evaluation.}
  \label{tab:validator_rules}
  \small
  \setlength{\tabcolsep}{3pt}
  \renewcommand{\arraystretch}{1.18}

  \begin{tabularx}{\linewidth}{
    >{\raggedright\arraybackslash}p{0.16\linewidth}
    >{\raggedright\arraybackslash}X
    >{\raggedright\arraybackslash}X
    >{\raggedright\arraybackslash}p{0.18\linewidth}
  }
    \toprule
    \makecell[l]{S-CoT\\field}
    &
    \makecell[l]{Trigger\\condition}
    &
    \makecell[l]{Behavioral\\constraint}
    &
    \makecell[l]{Violation\\type}
    \\
    \midrule

    Critical objects
    &
    Immediate obstacle on the ego path
    &
    The longitudinal action must be
    \texttt{decelerate} or \texttt{stop}
    &
    Hazard ignorance
    \\

    Environmental condition
    &
    Red light or stop sign affecting the ego lane
    &
    The longitudinal action must be
    \texttt{stop}
    &
    Right-of-way violation
    \\

    Critical objects
    &
    Pedestrian or vulnerable road user on the ego path
    &
    The longitudinal action must be
    \texttt{decelerate} or \texttt{stop}
    &
    Vulnerable-user risk
    \\

    Critical objects
    &
    Close leading or cut-in vehicle ahead
    &
    \texttt{accelerate} is forbidden
    &
    Following-risk violation
    \\

    Lane availability
    &
    Target lane is blocked or unavailable
    &
    A lane change toward the blocked side is forbidden
    &
    Invalid lane commitment
    \\

    Lane availability and route
    &
    The route requires a transition to an available target lane
    &
    The lateral action must change toward the target lane
    &
    Missed route-lane response
    \\

    Ambiguity label
    &
    A critical object, lane state, or visibility condition is uncertain
    &
    \texttt{accelerate} is forbidden
    &
    Overconfident reasoning
    \\

    Critical objects and lanes
    &
    No active signal constraint or critical object exists, and the current lane is free
    &
    An unexplained \texttt{stop} or
    \texttt{decelerate} is penalized
    &
    Phantom braking
    \\

    Reasoning intention and meta-action
    &
    The rationale contains constrained lateral or longitudinal intention verbs
    &
    The final meta-action must be compatible with the stated intention
    &
    Intention--action inconsistency
    \\

    \bottomrule
  \end{tabularx}
\end{table*}

\textbf{Policy Optimization.} During GRPO, given a visual input $q$ and a policy context $c$ determined by the frozen Arbiter, the VLM policy samples a group of candidate outputs $\{o_1, \dots, o_G\}$. The group-wise relative advantage $A_i$ is computed as:
\begin{align}
  A_i &= \frac{\mathcal{R}^i - \text{mean}(\{\mathcal{R}^j\}_{j=1}^G)}{\text{std}(\{\mathcal{R}^j\}_{j=1}^G) + \epsilon}.
\end{align}
The final objective uses a clipped loss with clipping parameter $\epsilon$ and a KL divergence penalty to prevent excessive deviation from the SFT reference model. It is formulated as follows:
\begin{equation}
\label{eq:grpo}
\begin{gathered}
  J(\theta) = \mathbb{E}_{q, c, \{o_i\}} \left[L_{\text{policy}}(\theta) - \beta \mathbb{D}_{\text{KL}}(\theta) \right], \\
  L_{\text{policy}}(\theta) = \sum_{i=1}^{G} \min \left( r_i(\theta) A_i, \text{clip}(r_i(\theta), 1-\epsilon, 1+\epsilon) A_i \right),
\end{gathered}
\end{equation}
where $L_{\text{policy}}$ denotes the policy optimization term, $\mathbb{D}_{\text{KL}}$ denotes the KL divergence regularization term, and $r_i(\theta) = \frac{\pi_{\theta}(o_i|q, c)}{\pi_{\theta_{\text{old}}}(o_i|q, c)}$ is the importance sampling ratio.


\section{Experiments}
\label{sec:experiments}

\subsection{Experimental Setup}
\textbf{Dataset.} 
Training and validation samples are drawn from disjoint driving logs in the NAVSIM trainval split, whereas evaluation samples are drawn from the test split \cite{dauner2024navsim}. The resulting dataset comprises 69,842 training samples (42,163 simple and 27,679 complex), 2,346 validation samples (1,407 simple and 939 complex), and 574 manually verified test samples (146 simple and 428 complex). Model checkpoints, the routing threshold, and reward weights are selected using the validation set, while the held-out test set is reserved for final evaluation. To probe behavior under selected out-of-distribution (OOD) and long-tail conditions, we construct external evaluation subsets containing 96--108 samples each. These consist of construction-zone cases from a public roadwork dataset \cite{ghosh2025roadwork}, traffic-sign reasoning cases from a public map-and-sign dataset \cite{chang2025driving}, and non-public real-world cases covering night driving, adverse weather, complex intersections, and dense urban interactions. All external subsets are used exclusively for evaluation and are excluded from both SFT and RLVR training.

\textbf{Metrics.} Planning performance is quantified by exact-match accuracy for the joint pair of lateral and longitudinal meta-actions and by classwise F1 scores within each action dimension. Routing performance is quantified by routing F1 and the false-negative rate (FNR) on complex scenes. CoT accuracy is reported only for the manual audit of data-engine outputs on complex scenes. An audited output is considered correct only if its planning-relevant reasoning and both meta-actions are correct. $F1_{\text{per}}$ denotes the sample-averaged set-based F1 score between the predicted and reference sets of perception items. The Logical Consistency Score (LCS) is computed using the validator in Table~\ref{tab:validator_rules}:
\begin{align}
  LCS &= \frac{N_{\text{pass}}}{N_{\text{complex}}},
\end{align}
where $N_{\text{pass}}$ denotes the number of samples in the complex-scene evaluation set that pass all logical-consistency rules, and $N_{\text{complex}}$ denotes the total number of samples in that set. LCS therefore quantifies rule-based internal consistency between the parsed reasoning fields and final meta-actions within the complex-scene evaluation set.

\textbf{Implementation Details.} 
We use Qwen3-VL-4B as the VLM backbone. Each model input comprises a front-view image, ego-vehicle state variables, and navigation instructions. The learning rate is \(5\times10^{-6}\) for both supervised fine-tuning and GRPO-based alignment of the VLM backbone. The decoupled Arbiter is trained with a learning rate of \(1\times10^{-5}\), and its default routing threshold is $\tau=0.7$. All training stages of the proposed framework are run on eight NVIDIA H20 GPUs. Inference latency is measured with a batch size of 1 on a single NVIDIA H20 GPU using BF16 precision after model warm-up. Table~\ref{tab2} reports the mean per-sample latency across the 574-sample test set. For comparison, we evaluate zero-shot VLMs, SFT-only VLMs, AlphaDrive, a 2B variant of the proposed framework, and two lightweight non-reasoning planners: IDM+MOBIL \cite{treiber2000congested,kesting2007general} and a CIL-style behavior-cloning model adapted to the same discrete meta-action space \cite{codevilla2018end}.

\subsection{Data Engine Quality}
We compare the quality of the S-CoT annotations generated by the proposed data engine with those produced by a single-pass autoregressive (AR) pipeline and a modular-perception prompting pipeline. Because the perception modules run locally, the annotation-cost calculation includes only calls to the expert VLM. The data-engine stage that invokes the expert VLM requires approximately six calls per sample, corresponding to approximately 16 annotated samples per minute at the measured service throughput. This offline annotation throughput is distinct from the online inference latency reported in Table~\ref{tab2}.

For the manual audit, 195 complex scenes are sampled from the training set. Outputs produced by all three pipelines are anonymized and independently reviewed by two annotators. An output is judged correct only if its reasoning and both meta-actions are correct. To provide a limited validity check for the rule-based validator, the annotators separately assess reasoning--action consistency without considering factual correctness. The validator agrees with the resulting consensus consistency labels for 563 of the 585 outputs (96.2\%; Cohen's $\kappa=0.73$).

\begin{figure}[pos=t]
  \centering
  \includegraphics[width=\columnwidth]{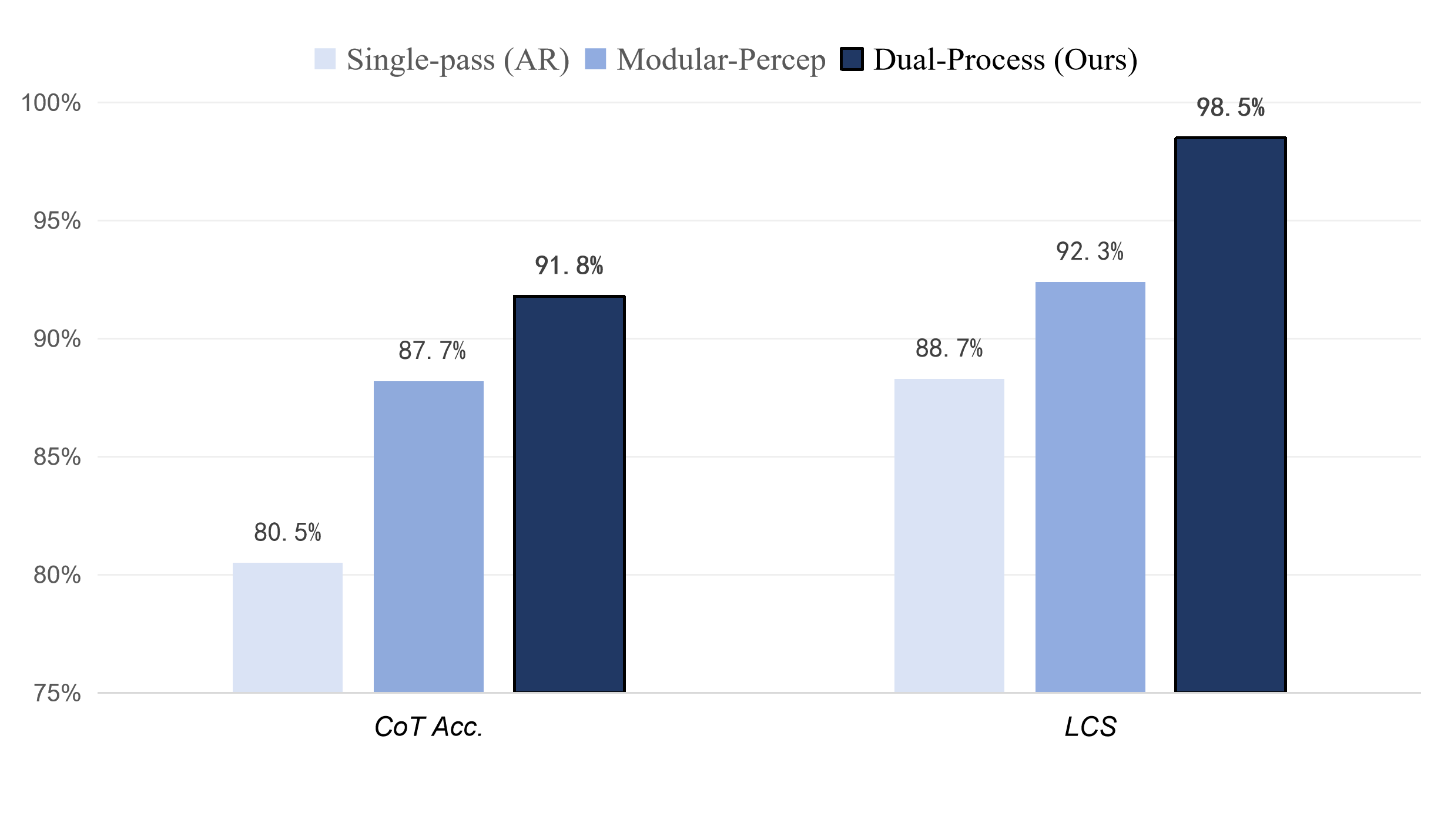}
  \caption{Chain-of-thought (CoT) accuracy and Logical Consistency Score (LCS) of structured chain-of-thought (S-CoT) annotations generated by the three data pipelines on the 195-scene audit subset.}
  \label{f3}
\end{figure}

Figure~\ref{f3} reports CoT accuracy and LCS for annotations generated by the three pipelines. The single-pass AR pipeline achieves a CoT accuracy of 80.5\% and an LCS of 88.7\%; these lower scores are consistent with its susceptibility to perception hallucinations. The modular-perception pipeline improves both metrics, but both remain below those achieved by the complete data engine. The complete data engine achieves 91.8\% CoT accuracy and 98.5\% LCS, the highest values among the three pipelines.

\subsection{Main Results}
Table~\ref{tab1} compares planning, perception, and logical-consistency performance on the NAVSIM test set. The rule-based IDM+MOBIL and CIL-BC baselines serve as lightweight planning-only references but yield lower F1 scores, particularly for non-default lateral and longitudinal maneuvers. The zero-shot VLMs achieve exact-match planning accuracies of only 13.07\% and 17.77\%, indicating limited transfer without task-specific fine-tuning. SFT substantially improves planning accuracy for both VLM backbones; however, LCS values of 66.82\% and 62.15\% for the corresponding SFT models indicate that supervised training alone does not reliably enforce reasoning--action consistency.

\begin{figure}[pos=t]
  \centering
  \includegraphics[width=\columnwidth]{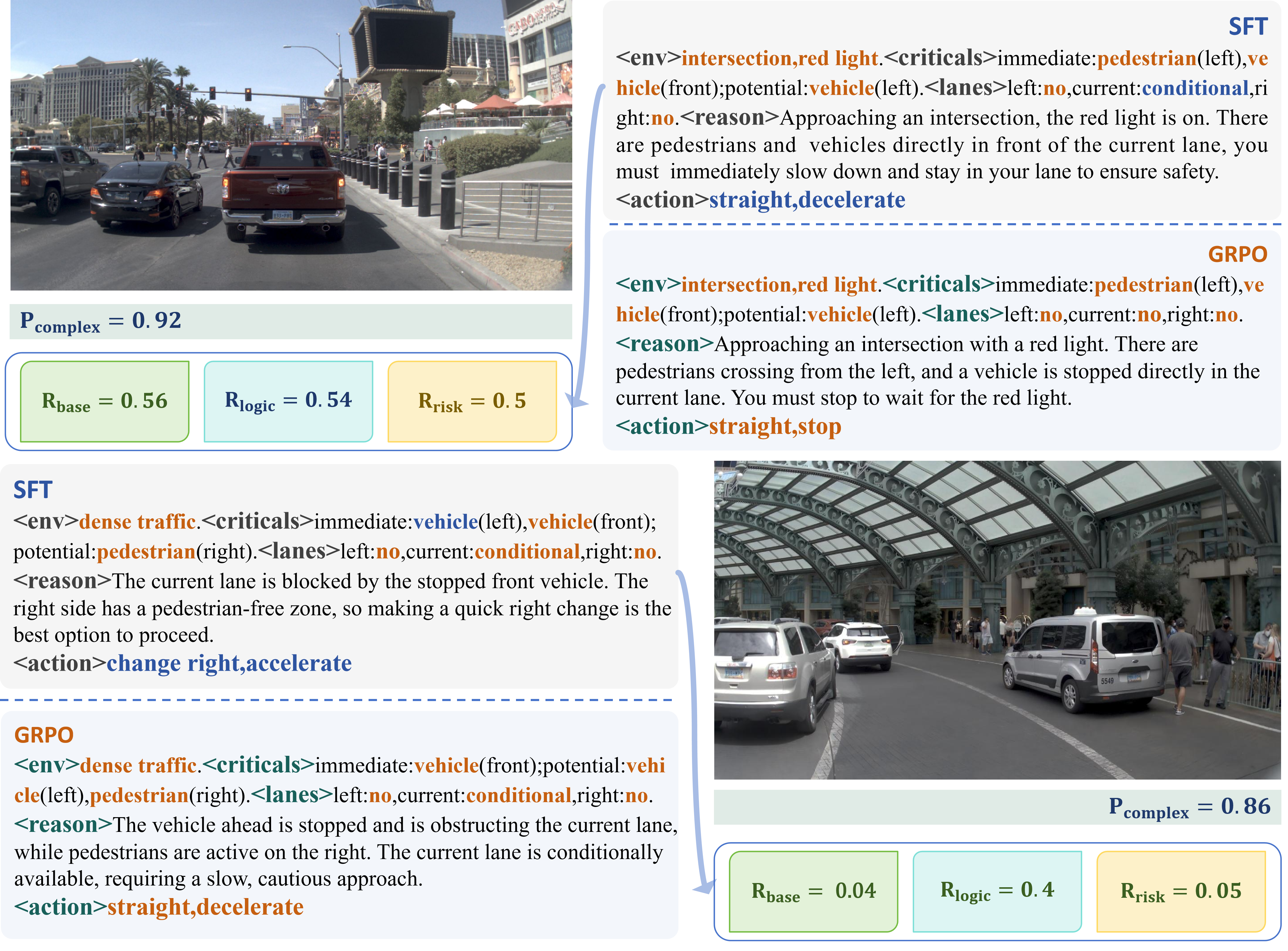}
  \caption{Qualitative comparison of reasoning--action consistency in a complex scene. The supervised fine-tuning (SFT) baseline produces an action that contradicts the perceived hazard, whereas the Group Relative Policy Optimization (GRPO)-aligned policy produces an action consistent with its stated reasoning.}
  \label{f4}
\end{figure}

The verifiable reinforcement-learning stage strengthens this alignment. Relative to the Qwen3-VL SFT baseline, the GRPO-aligned policy increases LCS by 35.05 percentage points, from 62.15\% to 97.20\%, while increasing planning accuracy from 70.73\% to 80.14\% and $F1_{\text{per}}$ from 78.62\% to 88.46\%. Figure~\ref{f4} provides a qualitative example: in the illustrated complex scene, the SFT model produces an action that contradicts the perceived hazard, whereas the GRPO-aligned policy produces an action consistent with its stated reasoning.

\begin{figure}[pos=t]
  \centering
  \includegraphics[width=\columnwidth]{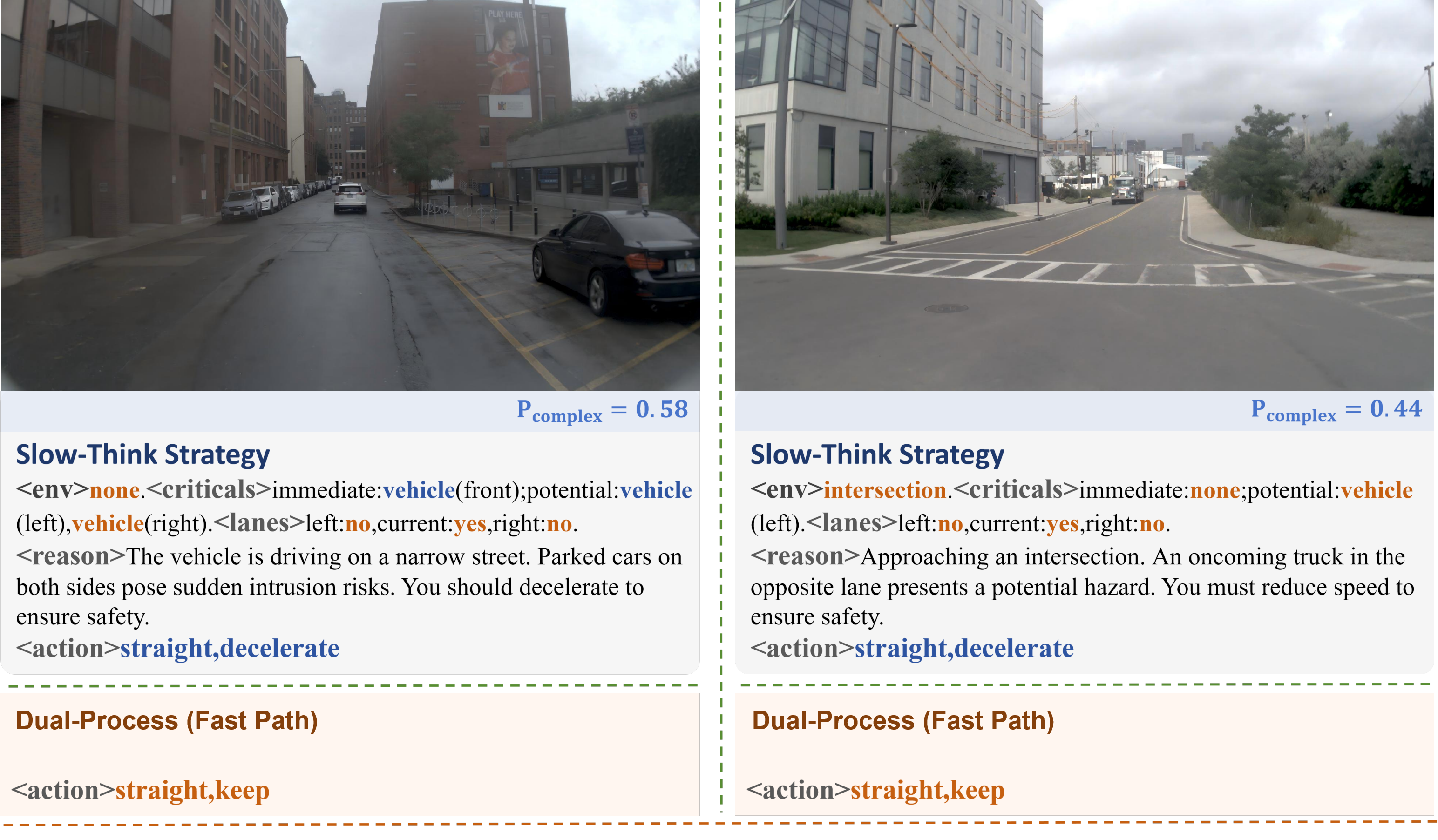}
  \caption{Qualitative example of over-reasoning in a simple scene. In this example, full structured reasoning produces a hallucinated risk and an unnecessary conservative maneuver, whereas dynamic routing avoids invoking the slow reasoning path.}
  \label{f5}
\end{figure}

\begin{table*}[pos=tbp]
  \centering
  \caption{
    Planning, perception, and Logical Consistency Score (LCS) results on the manually verified 574-sample NAVSIM test set. All reported performance metrics are percentages. $^{\dagger}$ denotes a model trained only by supervised fine-tuning (SFT).
  }
  \label{tab1}
  \resizebox{\textwidth}{!}{%
  \begin{tabular}{l|c|c|ccc|cccc|cc}
    \toprule

    \multirow{2}{*}{Method} & \multirow{2}{*}{Size} & \multirow{2}{*}{Plan. Acc.} & \multicolumn{3}{c|}{Lateral (F1)} & \multicolumn{4}{c|}{Longitudinal (F1)} & \multirow{2}{*}{$F1_{\text{per}}$} & \multirow{2}{*}{LCS} \\

    & & & straight & left & right & keep & accel. & decel. & stop & & \\

    \midrule
    InternVL3.5 & 4B & 13.07 & 27.25 & 17.48 & 15.33 & 47.88 & 4.42 & 12.74 & 9.54 & 34.61 & 74.30 \\
    Qwen3-VL & 4B & 17.77 & 33.64 & 18.74 & 16.42 & 42.34 & 11.43 & 17.53 & 12.27 & 43.52 & 72.20 \\
    \midrule
    IDM+MOBIL & Rule & 58.36 & 81.46 & 54.82 & 56.13 & 70.28 & 24.71 & 43.96 & 51.34 & - & - \\
    CIL-BC & 25M & 63.76 & 84.73 & 68.14 & 69.02 & 73.65 & 35.20 & 52.48 & 60.13 & - & - \\
    \midrule
    InternVL3.5$^{\dagger}$ & 4B & 66.38 & 85.19 & 76.64 & 78.42 & 77.25 & 41.66 & 63.72 & 72.24 & 72.37 & 66.82 \\
    Qwen3-VL$^{\dagger}$ & 4B & 70.73 & 88.98 & 86.34 & 84.75 & 77.14 & 46.05 & 70.74 & 75.83 & 78.62 & 62.15 \\
    AlphaDrive & 2B & {73.52} & {95.24} & {89.62} & {87.41} & {82.42} & {60.33} & {70.41} & \textbf{83.54} & - & - \\
    Dual-Process (2B) & 2B & {77.35} & {95.57} & {91.23} & {88.85} & {85.29} & {64.02} & {72.44} & {82.06} & {83.37} & {96.73} \\
    \textbf{Dual-Process (Ours)} & 4B & \textbf{80.14} & \textbf{96.35} & \textbf{93.71} & \textbf{91.41} & \textbf{87.67} & \textbf{67.26} & \textbf{76.22} & {83.43} & \textbf{88.46} & \textbf{97.20} \\

    \bottomrule

  \end{tabular}
  }
\end{table*}

\subsection{Reasoning Efficiency and Over-Reasoning Mitigation}
As shown in Table~\ref{tab2}, the Dual-Process model improves planning accuracy by 6.27 percentage points relative to static fast-path inference (Fast-Plan), at the cost of a 7.02\% increase in latency. Relative to static slow-path inference with CoT enabled for every scene (Slow-Think), it improves planning accuracy by 2.27 percentage points while reducing mean latency by 17.39\%.

The lower accuracy of Slow-Think relative to Dual-Process (77.87\% versus 80.14\%), together with the qualitative example in Figure~\ref{f5}, suggests an over-reasoning effect. In unobstructed cruising scenes, a complete reasoning chain may redirect attention toward peripheral entities and elicit responses to weak or hallucinated risks. Selective routing may reduce this type of failure by reserving slow-path reasoning for scenes with stronger visual or semantic evidence of complexity.

\begin{table}[pos=tbp]
\centering
\caption{Inference latency, routing performance, and planning accuracy under four reasoning modes.}
\label{tab2}
\setlength{\tabcolsep}{3pt}
\begin{tabular}{l c c c} 
    \toprule 
    Method & Latency (ms) & Routing F1 & Plan. Acc. \\
    \midrule
    Fast-Plan & \textbf{401.62} & - & 73.87 \\ 
    Slow-Think & 520.26 & - & 77.87 \\ 
    Implicit-Routing & 445.25 & 82.41 & 78.57 \\
    \textbf{Dual-Process (Ours)} & 429.81 & \textbf{89.46} & \textbf{80.14} \\ 
    \bottomrule 
\end{tabular}
\end{table}

\subsection{Long-Tail Robustness and Failure Analysis}

We evaluate the framework on the external long-tail subsets defined in the experimental setup. As shown in Table~\ref{tab:ood_extreme}, the average routing F1, planning accuracy, and FNR across the six subsets are 75.55\%, 65.47\%, and 25.10\%, respectively. The average LCS decreases from 97.20\% on the NAVSIM test set to 85.18\% on the external subsets, indicating that rule-aligned consistency transfers only partially beyond the training distribution. The traffic-sign reasoning subset yields the lowest values, and the low-visibility subset the second-lowest values, for routing F1, planning accuracy, perception F1, and LCS; the other four subsets yield higher but heterogeneous results across these metrics.

\begin{table*}[pos=tbp]
  \centering
  \caption{Routing, planning, perception, and Logical Consistency Score (LCS) performance on six external out-of-distribution (OOD) and long-tail driving subsets.}
  \label{tab:ood_extreme}
  \resizebox{\textwidth}{!}{%
  \begin{tabular}{l|c|c|c|c|c}
    \toprule
    Evaluation subset & FNR (\%) $\downarrow$ & Routing F1 (\%) $\uparrow$ & Planning Acc. (\%) $\uparrow$ & $F1_{\text{per}}$ (\%) $\uparrow$ & LCS (\%) $\uparrow$ \\
    \midrule
    Night driving & 19.84 & 78.92 & 71.26 & 76.88 & 88.47 \\
    Low visibility in rain or snow & 33.76 & 69.84 & 59.72 & 68.41 & 80.92 \\
    Construction zones & 16.73 & 83.18 & 66.84 & 79.15 & 89.12 \\
    Complex intersections & 20.46 & 78.34 & 69.38 & 75.82 & 87.52 \\
    Dense urban interactions & 21.18 & 77.56 & 68.71 & 76.34 & 86.18 \\
    Traffic-sign reasoning scenes & 38.62 & 65.47 & 56.88 & 65.23 & 78.84 \\
    \midrule
    Average & 25.10 & 75.55 & 65.47 & 73.64 & 85.18 \\
    \bottomrule
  \end{tabular}
  }
\end{table*}

\begin{figure*}[pos=t]
  \centering
  \includegraphics[width=0.9\textwidth]{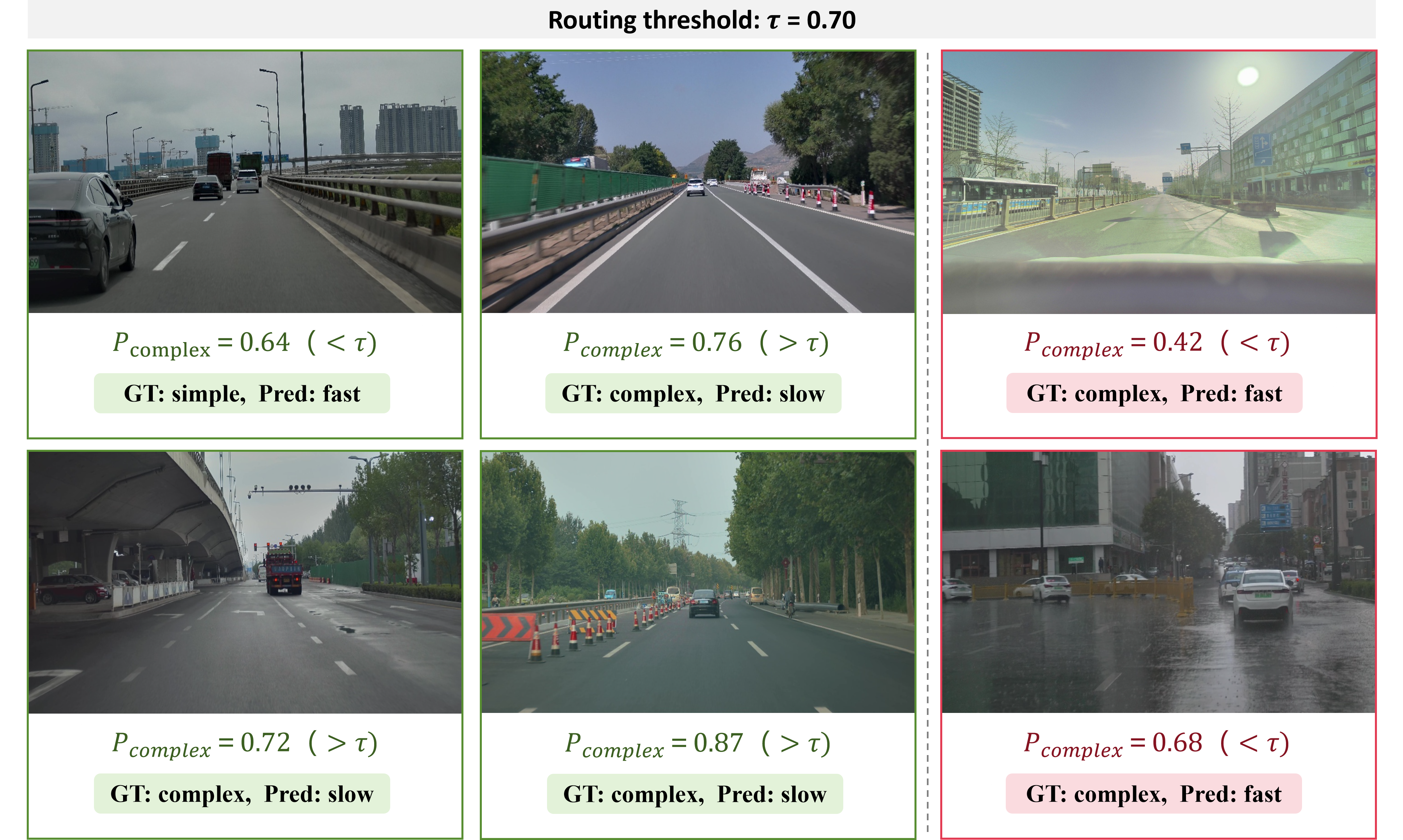}
  \caption{
    Qualitative routing analysis in out-of-distribution (OOD) and long-tail driving scenarios. The left panel shows correctly routed examples, with simple scenes assigned to the fast path and complex scenes assigned to the slow path. The right panel shows false-negative cases in which complex scenes receive low complexity scores and are incorrectly assigned to the fast path. With the routing threshold set to \(\tau=0.70\), samples satisfying \(P_{\text{complex}}\ge\tau\) are assigned to the slow path.
  }
  \label{fig:S1}
\end{figure*}

The false-negative examples in Figure~\ref{fig:S1} illustrate two failure modes. In low-visibility scenes involving rain or snow, visual degradation may obscure small obstacles, blur lane boundaries, and weaken the local cues available to the Arbiter. In traffic-sign reasoning scenes, relevant signs are often small, peripheral, or overlap only weakly with the ego vehicle's driving corridor; the current critical-path filter and Arbiter may therefore underestimate their planning relevance. These observations motivate future work on sign-aware grounding, feature calibration robust to adverse weather, and more explicit long-tail priors for routing.

\subsection{Ablation Studies and Sensitivity Analysis}
\textbf{Reward ablation.}
Table~\ref{tab4} isolates the incremental effects of the reward components used for GRPO. Relative to configuration 1, adding $\mathcal{R}_{\text{logic}}$ in configuration 2 increases LCS from 71.50\% to 96.96\% and $F1_{\text{per}}$ from 77.68\% to 90.34\%, identifying it as the primary contributor to the observed consistency gain. Adding $\mathcal{R}_{\text{risk}}$ in configuration 3 further increases planning accuracy from 79.27\% to 80.14\% and LCS from 96.96\% to 97.20\%, although $F1_{\text{per}}$ decreases from 90.34\% to 88.46\%. The risk-aversion reward addresses an early-training observation that cross-entropy loss and planning performance can diverge in ambiguous scenes, where the policy may become overconfident. It targets this behavior by rewarding explicit uncertainty and penalizing aggressive maneuvers under ambiguity. Thus, the complete reward configuration achieves the highest planning accuracy and LCS, with a modest reduction in perception F1 relative to configuration 2.

\begin{table*}[pos=tbp]
  \centering
  \caption{Effects of the verifiable reward components used in Group Relative Policy Optimization (GRPO) on planning, perception, and Logical Consistency Score (LCS) metrics.}
  \label{tab4}
  \resizebox{\textwidth}{!}{%
  \begin{tabular}{l|ccc|ccc|cccc|cc|c}
    \toprule
    \multirow{2}{*}{ID} & \multirow{2}{*}{$\mathcal{R}_{\text{base}}$} & \multirow{2}{*}{$\mathcal{R}_{\text{logic}}$} & \multirow{2}{*}{$\mathcal{R}_{\text{risk}}$} & \multicolumn{3}{c|}{Lateral (F1)} & \multicolumn{4}{c|}{Longitudinal (F1)} & \multirow{2}{*}{$F1_{\text{per}}$} & \multirow{2}{*}{LCS} & \multirow{2}{*}{Plan. Acc.} \\
    & & & & straight & left & right & keep & accel. & decel. & stop & & & \\
    \midrule
    1 & \checkmark & & & 90.46 & 92.41 & 89.43 & 83.58 & 60.78 & 70.54 & 72.66 & 77.68 & 71.50 & 75.61 \\
    2 & \checkmark & \checkmark & & 94.38 & \textbf{94.12} & 90.86 & 86.42 & 65.63 & 75.34 & 82.26 & \textbf{90.34} & 96.96 & 79.27 \\
    3 & \checkmark & \checkmark & \checkmark & \textbf{96.35} & 93.71 & \textbf{91.41} & \textbf{87.67} & \textbf{67.26} & \textbf{76.22} & \textbf{83.43} & 88.46 & \textbf{97.20} & \textbf{80.14} \\
    \bottomrule
  \end{tabular}%
  }
\end{table*}

\textbf{Training paradigm.}
Table~\ref{tab:tab_s1} compares the training paradigms. The SFT-only model reaches 70.73\% planning accuracy but 62.15\% LCS, whereas RL-only training reaches 85.75\% LCS but only 42.51\% planning accuracy without a supervised initialization. Although using an LLM-as-a-Judge improves upon pure SFT, its probabilistic reward signals introduce evaluation noise and less directly enforce driving-specific logical constraints. Combining SFT initialization with deterministic RLVR yields the highest Acc., $F1_{\text{per}}$, and LCS in this comparison.

\begin{table}[pos=tbp]
    \centering
    \caption{Comparison of supervised fine-tuning (SFT), reinforcement learning (RL)-only training, large-language-model (LLM)-judged RL, and deterministic Reinforcement Learning with Verifiable Rewards (RLVR).}
    \label{tab:tab_s1}
    \setlength{\tabcolsep}{5.5pt}
    \begin{tabular}{lccc}
        \toprule
        Paradigm & Planning Acc. (\%) & $F1_{\text{per}}$ (\%) & LCS (\%) \\
        \midrule
        $\pi_{\mathit{SFT}}$ & $70.73$ & $78.62$ & $62.15$ \\
        $\pi_{\mathit{RL\text{-}only}}$ & $42.51$ & $62.31$ & $85.75$ \\
        $\pi_{\mathit{SFT}\to\mathit{RL\text{-}LLM}}$ & $77.18$ & $81.33$ & $90.65$ \\
        $\boldsymbol{\pi_{\mathit{SFT}\to\mathit{RLVR}}}$ & $\mathbf{80.14}$ & $\mathbf{88.46}$ & $\mathbf{97.20}$ \\
        \bottomrule
    \end{tabular}
\end{table}

\begin{figure*}[pos=t]
  \centering
  \includegraphics[width=\textwidth]{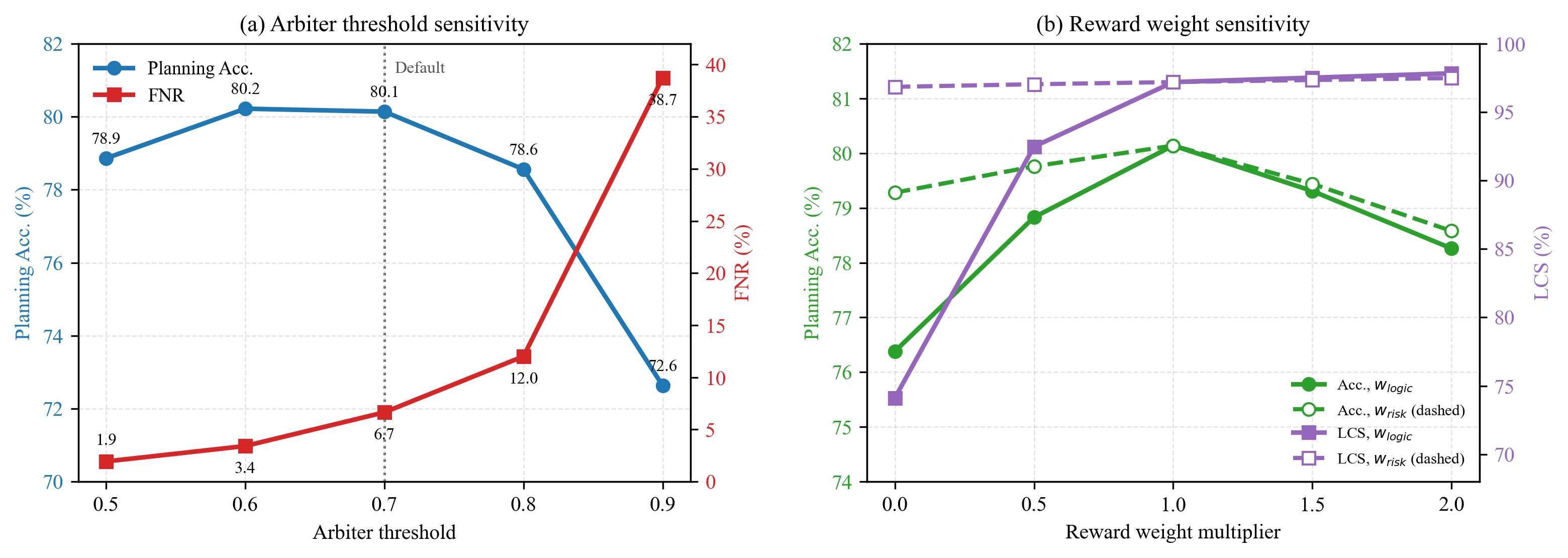}
  \caption{Sensitivity to the Arbiter threshold and verifiable-reward weights. The left panel reports planning accuracy and false-negative rate (FNR) at five thresholds; the dotted line marks the default setting (\(\tau=0.7\)), selected based on the joint trade-off among planning accuracy, FNR, routing F1, and latency. In the right panel, each verifiable-reward weight is varied while the other remains at its default value of 1.0 and the base-task reward is held fixed.}
  \label{fig:sensitivity}
\end{figure*}

\textbf{Sensitivity analysis.}
Figure~\ref{fig:sensitivity} examines sensitivity to the Arbiter threshold and the two verifiable-reward weights. 
At \(\tau=0.6\), planning accuracy is marginally higher than at \(\tau=0.7\) (80.22\% versus 80.14\%) and FNR is lower (3.42\% versus 6.65\%). However, routing F1 decreases from 89.46\% to 72.35\%, and latency increases from 429.81 to 471.48 ms, indicating excessive use of the slow path. At \(\tau=0.8\), latency decreases to 415.74 ms, but routing F1 and planning accuracy decrease to 76.28\% and 78.56\%, respectively, while FNR increases to 12.02\%. We therefore retain \(\tau=0.7\) as a practical operating point that balances routing quality, planning accuracy, and inference cost.
The reward-weight analysis indicates that \(\mathcal{R}_{\text{logic}}\) is the primary driver of LCS improvement, whereas \(\mathcal{R}_{\text{risk}}\) has a smaller but consistent effect. Increasing either weight beyond 1.0 yields only limited additional LCS gains and slightly reduces planning accuracy, suggesting that overweighting the verification terms can induce overly conservative behavior.

\FloatBarrier


\section{Conclusion}
\label{sec:conclusion}

This study presents a cognitive dual-process framework for integrating structured scene knowledge and adaptive reasoning into the knowledge-intensive task of high-level VLM planning. The framework combines an automated data engine that constructs machine-parsable S-CoT supervision, a pre-decoding visual Arbiter that routes scenes between direct planning and structured reasoning, and deterministic verification that supplies training signals for consistency between parsed scene factors and discrete meta-actions. In the manual audit, the generated annotations achieve 91.8\% CoT accuracy and 98.5\% LCS. On the manually verified 574-sample NAVSIM test set, the planner achieves 80.14\% planning accuracy and 97.20\% LCS while reducing mean latency by 17.39\% relative to static slow reasoning. Results on the external long-tail subsets show partial transfer beyond the training distribution but also identify the largest performance degradations in low-visibility and traffic-sign reasoning scenes. Taken together, the results indicate that adaptive routing can reduce redundant VLM inference, while explicit scene-knowledge representation and rule-based verification support planning performance and reasoning--action consistency within the evaluated high-level planning setting.

Several limitations remain. Although the Arbiter produces a continuous complexity score, the visual evidence supporting an individual routing decision is not yet directly interpretable. The current S-CoT schema and validator cover a fixed set of scene factors and may not capture uncommon interactions, particularly under severe visual degradation. Moreover, the external evaluation uses targeted subsets rather than a comprehensive cross-dataset benchmark; broader generalization therefore remains to be established. Finally, the planner predicts discrete meta-actions rather than executable trajectories, so the present results do not establish closed-loop driving performance. Future work will investigate routing attribution and calibration, extend the S-CoT schema and validator to additional long-tail conditions, and integrate high-level reasoning with trajectory generation and closed-loop evaluation.












\printcredits


\bibliographystyle{cas-model2-names}

\bibliography{main}



\end{document}